\documentclass[10pt,twocolumn,letterpaper]{article}

\usepackage{cvpr}      

\usepackage{graphicx}
\usepackage{amsmath}
\usepackage{amssymb}
\usepackage{booktabs}
\usepackage{diagbox}

\usepackage[pagebackref,breaklinks,colorlinks]{hyperref}

\usepackage{algorithm}
\usepackage{algorithmic}

\usepackage{amsmath}
\usepackage{amssymb}

\newcommand\blfootnote[1]{%
\begingroup
\renewcommand\thefootnote{}\footnote{#1}%
\addtocounter{footnote}{-1}%
\endgroup
}

\usepackage{color}
\usepackage{subfigure}
\usepackage[switch]{lineno}
\newcommand*{\affaddr}[1]{#1} 
\newcommand*{\affmark}[1][*]{\textsuperscript{#1}}

\usepackage{hyperref}
\usepackage{marvosym}

\usepackage{caption}
\usepackage[capitalize]{cleveref}
\crefname{section}{Sec.}{Secs.}
\Crefname{section}{Section}{Sections}
\Crefname{table}{Table}{Tables}
\crefname{table}{Tab.}{Tabs.}

\begin{document}

\title{FaceChain: A Playground for Human-centric Artificial Intelligence Generated Content}

\author{
Yang Liu \affmark[1]\textsuperscript{*},
Cheng Yu \affmark[1]\textsuperscript{*},
Lei Shang \affmark[1],
Yongyi He \affmark[1],
Ziheng Wu \affmark[1],
Xingjun Wang \affmark[1],
Chao Xu \affmark[1], \\
Haoyu Xie \affmark[1], 
Weida Wang \affmark[2],
Yuze Zhao  \affmark[1], 
Lin Zhu \affmark[1], 
Chen Cheng \affmark[1], 
Weitao Chen \affmark[1],
Yuan Yao \affmark[1], \\
Wenmeng Zhou \affmark[1], 
Jiaqi Xu \affmark[1],
Qiang Wang \affmark[1],
Yingda Chen \affmark[1] \textrm{\Letter} ,
Xuansong Xie \affmark[1] \textrm{\Letter},
Baigui Sun  \affmark[1] \textrm{\Letter} \\
\affaddr{\affmark[1]Alibaba Group} \qquad \affaddr{\affmark[2]FaceChain Community} \\
\{ly261666, yucheng.yu, sl172005, yongyi.hyy, zhoulou.wzh, xingjun.wxj, xc264362, xiehaoyu.xhy, \\
yuze.zyz, lin.zhu, chengchen.cc, 
weitao.cwt, ryan.yy, wenmeng.zwm,
zhoumo.xjq, yongyi.hyy \\
yingda.chen, xingtong.xxs, baigui.sbg\}@alibaba-inc.com \\
}

\twocolumn[{
    \renewcommand\twocolumn[1][]{#1}
    \maketitle
    \begin{center}
    \includegraphics[width=0.85\linewidth]{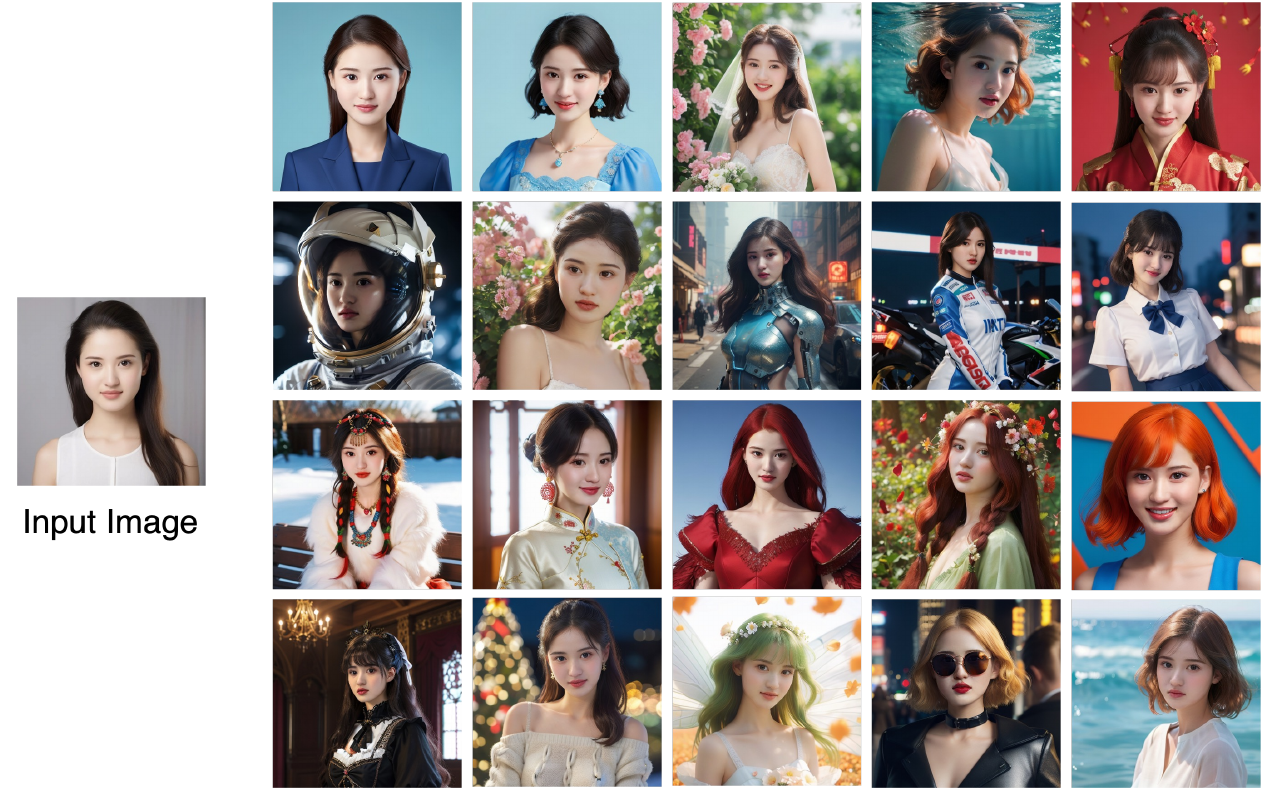}
    \captionsetup{type=figure}
    \caption{FaceChain Generation Results}
    \label{fig:infinite_style_1}
\end{center}
}]

\maketitle

\textbf{\textit{Abstract:}} Recent advancement in personalized image generation have unveiled the intriguing capability of pre-trained text-to-image models on learning identity information from a collection of portrait images.
However, existing solutions are vulnerable in producing truthful details, and usually suffer from several defects such as
(i) The generated face exhibit its own unique characteristics, \ie facial shape and facial feature positioning may not resemble key characteristics of the input, and
(ii) The synthesized face may contain warped, blurred or corrupted regions.
In this paper, we present FaceChain, a personalized portrait generation framework that combines a series of customized image-generation model and a rich set of face-related perceptual understanding models (\eg, face detection, deep face embedding extraction, and facial attribute recognition), to tackle aforementioned challenges and to generate truthful personalized portraits, with only a handful of portrait images as input.
Concretely, we inject several SOTA face models into the generation procedure, achieving a more efficient label-tagging, data-processing, and model post-processing compared to previous solutions, such as DreamBooth ~\cite{ruiz2023dreambooth} , InstantBooth ~\cite{shi2023instantbooth} , or other LoRA-only approaches ~\cite{hu2021lora} .
We have designed FaceChain as a framework comprised of pluggable components that can be easily adjusted to accommodate different styles and personalized needs.
Besides, based on FaceChain, we further develop several applications to build a broader playground for better showing its value, including virtual try on and 2D talking head.
We hope it can grow to serve the burgeoning needs from the communities. 
Note that this is an ongoing work that will be consistently refined and improved upon.
FaceChain is open-sourced under Apache-2.0 license at \url{https://github.com/modelscope/facechain}.
\blfootnote{
\textsuperscript{*} Equal Contribution, \textrm{\Letter} Corresponding Author \\
}

\begin{figure*}[t]
    \centering
    \includegraphics[width=0.95\linewidth]{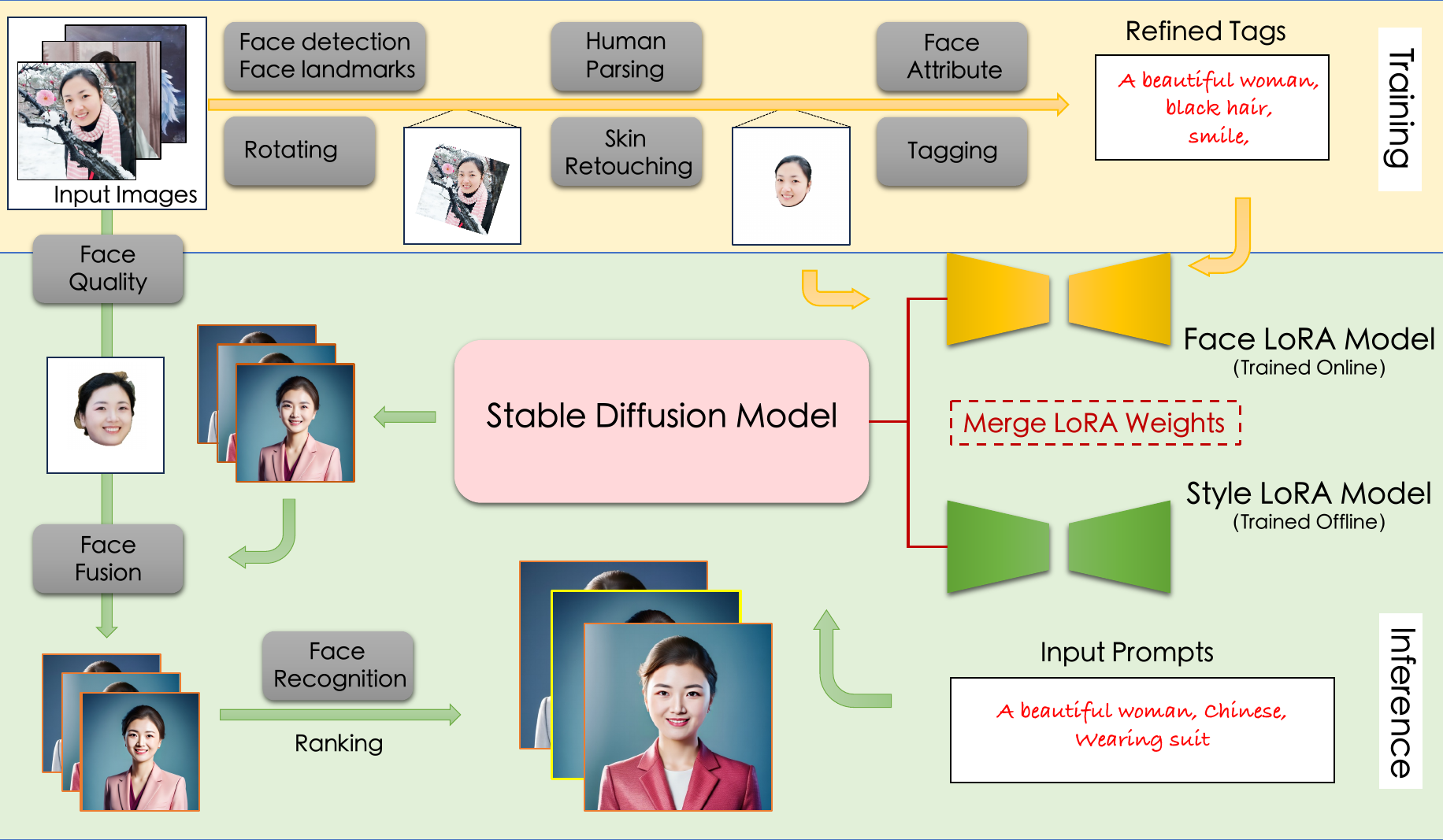}
    \caption{\textbf{Architectural overview of FaceChain personalized portrait generation.} 
    During training, multiple data processing approaches are adopted to generate tagged face images to train face-LoRA model online. 
    The weights of face-LoRA and style-LoRA models are then merged into foundation Stable-Diffusion model for text-to-image generation. 
    The generated portraits go through post-face-fusion and ranking before returning to users.
    }
    \label{fig:overview}
\end{figure*}

\section{Introduction}
\label{sec:introduction}

Recent years have witnessed a remarkable progress in the field of text-to-image generation, with large models ~\cite{rombach2022high, leedalle} emerging as the powerful foundation for creating high-fidelity and diverse images.
Given a text prompt, these models have demonstrated the impressive ability to create realistic and detailed image, showcasing their potential for a wide range of applications, \eg content generation, virtual reality and augmented reality.
However, for human-centric content generation, pre-trained text-to-image models often struggle to produce satisfactory portrait images that retain identities of individuals. This can be frustrating for individuals who wish to generate self-portraits. The imperfection arises due to the inherent limitations of these models, which are not designed to accurately preserve identity information. To this end, several recent efforts have started with the emphasis on tackling faithful personalized text-to-image generation. These efforts aim to learn the identity information from a collection of portrait images, then generate new scenes or styles corresponding to the target human beings under the guidance of text prompt.

The existing human-centric personalized text-to-image generation methods can be categorized into 
(i) LoRA-based Framework, which leverages the LoRA (Low-Rank Adaptation) fine-tune ~\cite{hu2021lora} technology on text-to-image model (\eg Stable Diffusion ~\cite{rombach2022high}) to generate identity-preserved images.
(ii) Identifier-based methods \cite{ruiz2023dreambooth, shi2023instantbooth}, which aim to learn a unique identifier relevant to identity information.
Although these methods can synthesize identity-similar images, they still suffer from several defects, \eg, facial shape and facial feature positioning may differ significantly from the input face; the synthesized face may contain several warped and corrupted regions.
In this paper we present FaceChain, a identifying preserving framework that not only preserves the distinguishing features of faces but also allows for versatile control over stylistic elements. 
Specifically, we inject two LoRA models into Stable Diffusion model, imparting it with the ability to integrate personalized style and identity information simultaneously. In particular, FaceChain is rooted in the ModelScope (\url{https://modelscope.cn}), an open-source community that seeks to bring together models from different areas and offer them via a unified interface. This allows FaceChain to integrate a comprehensive suite of face-related models, in addition the foundation model, to build the framework that generate identity-preserving portraits, the process of which we detail later in Sec. ~\ref{sec:architecture}.
Then, in Sec. ~\ref{sec:application}, we further develop several applications to build a broader playground for advancing the development of Face/Human-centric AIGC. 
Sec. ~\ref{sec:future work} present a thorough discussion on how future research and applications can stem from and flourish on FaceChain. As a relatively new open-source project, we believe FaceChain has shown its potentials. Other than the practical functionalities, we also aspire to expanding it into the benchmark and playground, that inspires innovations in personalized text-to-image generation.


\section{Architecture}
\label{sec:architecture}
FaceChain encapsulates the process of personalized portrait generation within an atomic pipeline, and is built upon Stable Diffusion~\cite{rombach2022high} model.
To improve the style stability and ID consistency of text-to-image generation, we adopt LoRA~\cite{hu2021lora}, a parameter-efficient strategy to fine-tune Stable Diffusion model.
With the \textit{composability} of multiple LoRA models, we learn the information for portrait style and human identities with different LoRA models, namely the style-LoRA model and face-LoRA model respectively. These two models are trained separately via text-to-image training on images of given style and human identities.
Specifically, we choose to train the style-LoRA model offline, which we describe in Section~\ref{sec:model-training}, while the face-LoRA model is trained online using the images uploaded by users -- which are of the same human identity.
Since the quality of the user-uploaded images may vary, FaceChain incorporates a rich set of face-related perceptual understanding models to ensure face images feeding into the training process are normalized to meet certain quality standards, such as appropriate size, good skin quality,  correct orientation, as well as having accurate tags.
The weights of multiple LoRA models are then merged into the Stable Diffusion model during inference to generate personalized portraits.
Finally, the details of the generated portraits are further enhanced by a series of post-processing steps.
The overall processing pipeline is illustrated in Fig.~\ref{fig:overview}.
Furturmore, FaceChain also introuduces an alternative inpainting pipeline in Sec. ~\ref{sec:inpainting}, which allows the users to generate corresponding portrait photos with high ID fidelity. 
Finally, in Sec. ~\ref{sec:infinite-style}, we integrate more than 100 style-LoRA model into FaceChain for generating diverse portraits.

\subsection{Data Processing}
\subsubsection{Face Extraction}
To improve the training-stability for face-LoRA models, FaceChain chains a series of face-related data processing modules to extract faces with appropriate size, good skin quality, and correct orientation from the images uploaded by users. These modules leverage extensively, the various models available on ModelScope, which we list below.

\noindent \textbf{Image Rotation.}
The orientation of human in images uploaded by users may not be suitable for training. To rectify this, orientation of the uploaded image is first determined and image rotation is perform if necessary. First, a rotation angle determination model is adopted to predict the probabilities of the image rotation with angle $0^{\circ}$,  $90^{\circ}$, $180^{\circ}$, and $270^{\circ}$. Then, the image is rotated by the angle with the largest probability. The rotation angle judging model is available at \url{https://modelscope.cn/models/Cherrytest/rot_bgr}.

\noindent \textbf{Face Rotation.}
After the initial image rotation, face orientation within the image may still fall short of the training requirement. Therefore, we tail it with a more accurate face rotation module, which perform the rotation according to the location of facial landmarks.
In particular, DamoFD~\cite{liu2022damofd}, a face detector using Network Architecture Search~\cite{zoph2016neural} (NAS) is used to obtain the detection result of five face landmarks.  Rotation matrix is then computed for the coordinates of detected landmarks and standard face template, using the least square method.
During this process, the two sets of coordinates are firstly normalized, by subtracting their mean and divided by their standard deviation, to eliminate the effect of translation and scaling.
If we denote the normalized coordinates of the detected landmarks and the face template as $\mathbf{P}_1, \mathbf{P}_2 \in \mathbb{R}^{5 \times 2}$, respectively.
the aim here is to minimize $\Vert \mathbf{R} \mathbf{P}_{1}^{\top} - \mathbf{P}_{2}^{\top} \Vert^2$, where $\mathbf{R}$ is the rotation matrix formulated as Eq.~\ref{eq:rot}.
\begin{equation}
    \label{eq:rot}
    \mathbf{R} = \begin{pmatrix}
 \text{cos} \theta & \text{-sin} \theta \\
 \text{sin} \theta & \text{cos} \theta \\
\end{pmatrix}
\end{equation}
As such, $\mathbf{R} = ((\mathbf{P}_{1}^{\top} \mathbf{P}_{1})^{-1} \mathbf{P}_{1}^{\top} \mathbf{P}_{2})^{\top}$, and we get the rotation angle $\theta = \text{arctan} (\mathbf{R}_{21} / \mathbf{R}_{22})$.
The image is then rotated accordingly, as illustrated in Fig.~\ref{fig:overview}.
The DamoFD model is available at \url{https://modelscope.cn/models/damo/cv_ddsar_face-detection_iclr23-damofd}.

\noindent \textbf{Face Region Crop and Segmentation.}
After image and face rotation, the face regions are then cropped out and masked from the input images. Using the DamoFD model, 
the bounding box of the face is determined, we then crop the image and adjust the size and position of the face.
As such, we keep the face centered horizontally, with its size between 0.35 and 0.45 times the whole image size.
Then we use Masked-attention Mask Transformer~\cite{cheng2022masked} model for human parsing (M2FP) to generate the mask of the head region, and perform segmentation accordingly.
The M2FP model is available at \url{https://modelscope.cn/models/damo/cv_resnet101_image-multiple-human-parsing}.

\noindent \textbf{Skin Retouching.}
Since the skin quality of the images uploaded by users may be unpromising, we leverage the skin retouching module to improve the skin quality of the face images.
In particular, the Adaptive Blend Pyramid Network~\cite{lei2022abpn} (ABPN) is used here for skin retouching, which is available at \url{https://modelscope.cn/models/damo/cv_unet_skin-retouching}.

\subsubsection{Label Tagging}
High-quality tagging is critical to facilitate text-to-image training. As such, we train face-LoRA model to learn the relationship between the face images and the generated tags.
This allows the Stable Diffusion model to generate images with corresponding features during inference, when prompting with suitable tags.
Therefore, the tags must be labeled appropriately to ensure that face-LoRA model can be triggered to produce stable output.
We identified three major requirements for label tagging:
\begin{itemize}
    \item Tags specific to a given image, such as facial expressions, jewelries and  accessories, should be labeled accurately to retain the relationship between salient image features and the corresponding tags.
    \item Tags bound to  human identities,  such as eyes, lips and ears, can be removed. Instead, the LoRA model can be relied on to generate such features without any prompt words.
    \item In general, using one tag to describe the overall characteristics of the human identity works quite well in practice. For example, we may use a man/woman/boy/girl, as the trigger word for all images. As such, when adding the trigger word into input prompts, the features of the human identity can be generated more easily.
\end{itemize}

FaceChain combine different approaches for label tagging to satisfy the above requirements.
First, the face images are fed into DeepDanbooru~\footnote{available at \url{https://github.com/KichangKim/DeepDanbooru}}, a text annotation model, to get the preliminary tags.
Then, we perform tag post-processing to choose tags corresponding to human identities, and remove them to meet the first two requirements.
Finally, we use FairFace~\cite{karkkainen2021fairface}, a face attribute model, to predict the probabilities of the gender and age attributes for each image, and use the overall result as the final prediction for the human identity.
Then we choose the trigger word according to the prediction of gender and age, as is shown in Table~\ref{tab:trigger_word}.
The FairFace model is available at \url{https://modelscope.cn/models/damo/cv_resnet34_face-attribute-recognition_fairface}.

\begin{table}[t]
    \centering
    \caption{Trigger words describing different gender and age.}
    \begin{tabular}{@{\extracolsep{\fill}}ccc}     
       \hline
       \diagbox{Age}{Gender} & Male & Female \\
       \hline
       $0\sim 20$ & a boy, children & a girl, children \\
       $20\sim 40$ & a handsome man & a beautiful woman \\
       $>40$ & a mature man & a mature woman \\
       \hline
    \end{tabular}
    \label{tab:trigger_word}
\end{table}

\subsection{Model Training}
\label{sec:model-training}
Style-LoRA model acts as the anchor-stone for producing stable styles of portraits. It is important for personal portrait generation since it set the guardrails for image generation model at large. The main style-LoRA model used with FaceChain targets personal portraits, and is trained with a large number of portrait-like images of the same style, such as ID photos. We share here the hyper-parameters used for training the LoRA model here for full disclosure. The rank of LoRA model is set to 32. Learning rate is set to 1e-4, and cosine with restarts schedule is deployed. The LoRA model is trained for 20 epochs to produce the final model. We deploy 8bit AdamW optimizer~\cite{dettmers2022bit} to save on training hardware.
As to the training of face-LoRA model, the user-uploaded images are firstly rotated based on the angle predicted by the image rotation model. It is then followd by the face alignment method based on face detection and keypoint output, which obtains images containing forward-looking faces. Next, we use the human body parsing model and the human portrait beautification model to obtain high-quality face training images. Afterwards, we use a face attribute model and a text annotation model, combined with tag post-processing methods, to generate fine-grained labels for training images. Finally, we use the above images and label data to fine-tune the Stable Diffusion model to obtain the face-LoRA model.

\begin{figure*}[t]
    \centering
    \includegraphics[width=0.95\linewidth]{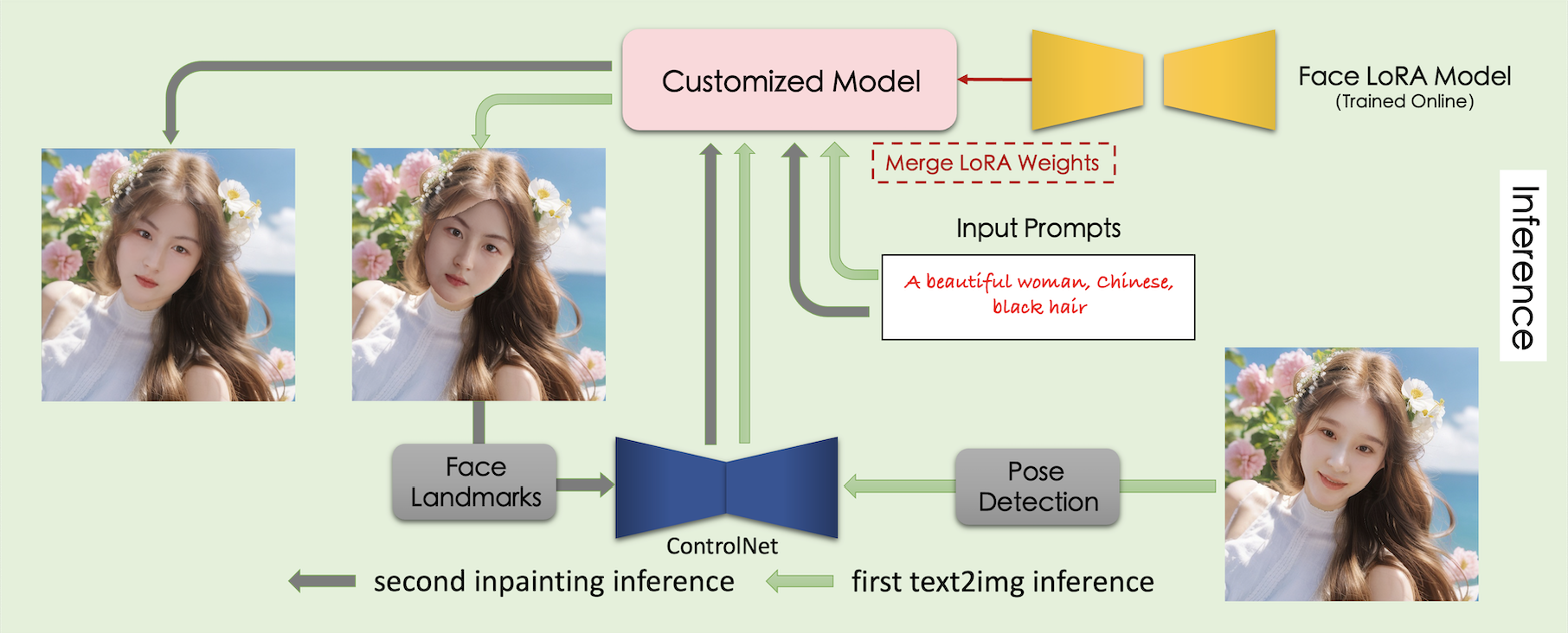}
    \caption{\textbf{Architectural overview of the inpainting pipeline of FaceChain personalized portrait generation.} 
    We first generate the face image through text-to-image inference guided by the bone pose of the template image. Then the warped face is used to extract the face landmarks to improve facial structure preservation. The final portraits are generated through the second inpainting inference.
    }
    \label{fig:overview_inpaint}
\end{figure*}

\subsection{Model Inference}
\label{sec:model-inference}
During inference phase, we fuse the weights of the face-LoRA model and style-LoRA model into the Stable Diffusion model. The fusing weights are chosen to be 0.25 and 1.0, respectively. Next, we use  Stable Diffusion‘s text-to-image generation pipeline to generate the preliminary personal portraits with preset input prompt words. Then we further improve the face details of the above generated portrait image and facial similarities with the input faces using a face fusion model. The template face used for fusion is selected from the training images via the face quality evaluation model. Finally, we use the face recognition model to calculate the similarity between the generated portrait image and the template face, the resulting portrait images are sorted and ranked accordingly before final output.

\subsection{Model Post Processing}
\label{sec:model-postpro}
After generating preliminary portraits by the Stable Diffusion model, FaceChain integrates several post process modules listed below to improve the face details and facial similarities of the portraits.
The overall pipeline for post process is illustrated in Fig.~\ref{fig:overview}.

\noindent \textbf{Template Face Selection.}
We adopt the Face Quality Assessment (FQA) model to evaluate the quality score for all faces from the user-uploaded images.
The face with the highest quality score is then chosen as the template-face for face fusion.
The FQA model is available at \url{https://www.modelscope.cn/models/damo/cv_manual_face-quality-assessment_fqa}.

\noindent \textbf{Face Fusion.}
We perform face fusion for the generated portraits using the selected template-face to improve facial details.
This allows the output portrait to retain major appearance features, while displaying more refined facial details.
The face fusion model is available at \url{https://www.modelscope.cn/models/damo/cv_unet-image-face-fusion_damo}.

\noindent \textbf{Similarity Ranking.}
The final output portraits are selected by comparing their facial similarities to the template face.
Given the inherent statistical difference between generated portraits and the input images, we adopt Random Temperature Scaling~\cite{shang2023improving} (RTS), a robust face recognition model for both in-distribution and out-of-distribution samples, to calculate the facial similarities .
Finally, the portraits with high facial similarity are selected as the output.
The RTS model is available at \url{https://www.modelscope.cn/models/damo/cv_ir_face-recognition-ood_rts}.

\subsection{Inpainting}
\label{sec:inpainting}
Utilizing a LoRA model specifically trained on a unique facial ID, coupled with an integrated stylistic background, enables the generation of relevant portrait photos with high ID fidelity. This approach also offers robust support for both the creativity and naturalness of the background.
Besides, in real-world scenarios and applications, there are often cases where users have already identified images they are satisfied with—images that feature well-designed compositions, lighting, and background arrangements. In such cases, the user's requirement is to naturally replace the face in the chosen image with a specific ID. 
To address this, we devise a framework based on the inpainting procedure~\cite{avrahami2022blended} of the Stable Diffusion model.
This allows for the natural redrawing of the designated facial ID on the basis of the specified template image.

\begin{figure*}[t]
    \centering
    \includegraphics[width=0.9\linewidth]{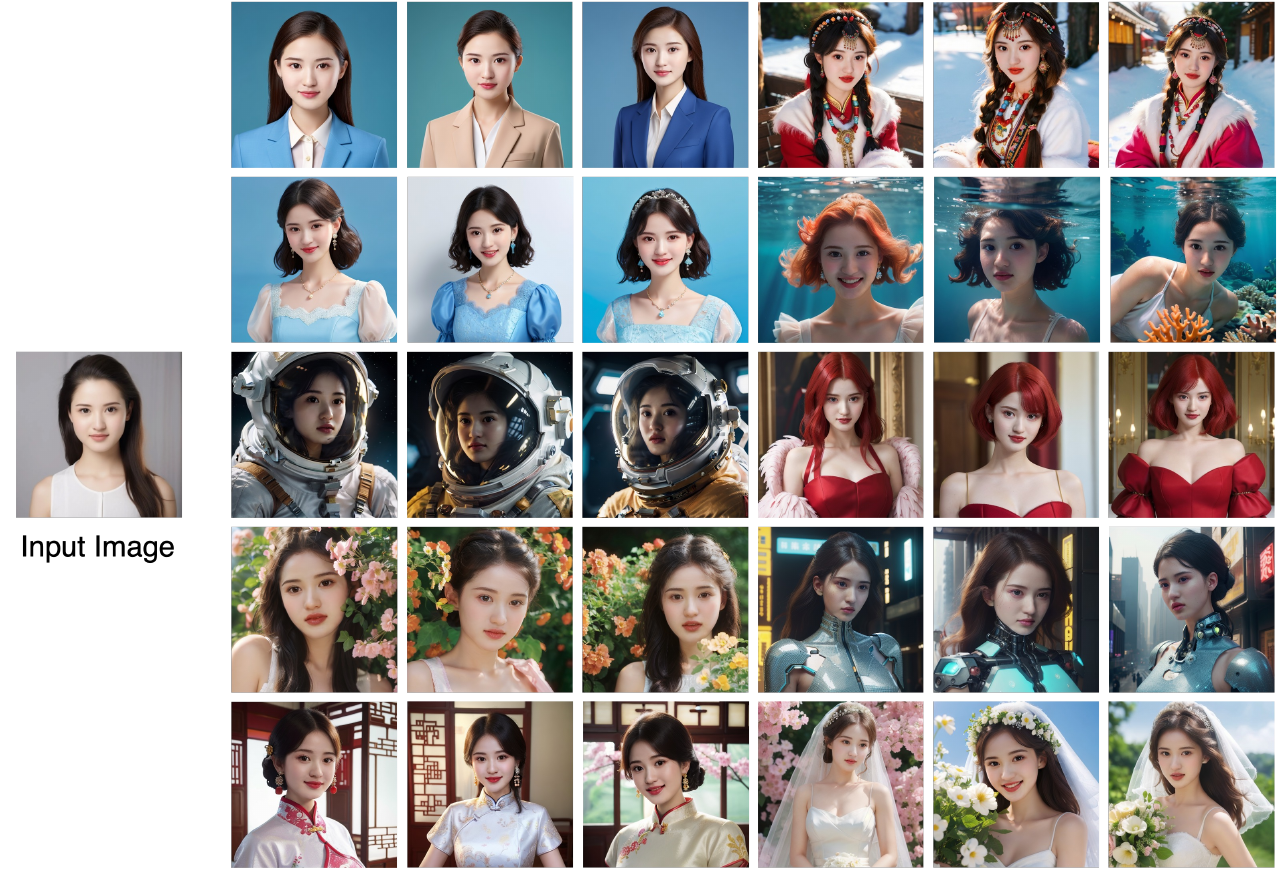}
    \caption{\textbf{Generated results with various style-LoRA models}
    }
    \label{fig:infinite_style_2}
\end{figure*}

The most straightforward solution is to perform diffusion mask inpainting on the face region of the input template with spatial conditional controls using ControlNet~\cite{zhang2023adding}, for which commonly used frameworks like OpenPose~\cite{cao2017realtime} and Canny edges~\cite{canny1986computational} can be employed. 
However, this direct approach to local redrawing introduces the following significant issues:
\begin{itemize}
    \item The facial landmarks in the original template are incompatible with those of the specified ID, leading to distortions in the facial structure and features when directly referencing the template's face landmarks or Canny edges.
    \item Mask inpainting incorporates different faces resulting in noticeable defects at the junctions between the redrawn areas and the original image, significantly compromising the overall realism of the photo.
    \item Combining multiple controls like ControlNet, LoRA models and image latent of the input template in the diffusion process leads to an adverse impact on each part, especially on identity preservation in face-LoRA models.
\end{itemize}

In order to solve the above problems, we propose a two-stage inpainting pipeline combining both text-to-image and inpainting procedures of the Stable Diffusion model as well as multiple ControlNets for natural redrawing of the face region.
Such a pipeline significantly improves the trade-off of the facial structure preservation of the specified ID and the overall realism at the junctions between the redrawn areas and the original image by progressively adding multiple controls in the diffusion process.
The overall inference pipeline is illustrated in Fig.~\ref{fig:overview_inpaint}.

\subsubsection{First Text-to-image Inference}
As is mentioned above, multiple controls such as the Canny edges and the image latent of the input template are required when inpainting the face region to guarantee the overall naturalness of the portrait, which compromises identity preservation, especially facial structure preservation in face-LoRA models.
Therefore, stronger facial structural control for the specified ID is needed.
To address this, we introduce a bone pose-guided text-to-image inference and face-warping algorithm to generate the 68 face landmarks as the input of OpenPose ControlNet in inpainting.
Only bone poses are used as the input of the OpenPose ControlNet in text-to-image inference, so as to strengthen the facial structure preservation of the face-LoRA model as far as possible, as well as maintaining the overall 3D orientation and position of the face.
Then, we warp the generated face to the input template to generate more accurate face landmarks.
Initially, we calculate the face landmarks $L_{template}$ and $L_{ID}$ of the features in the template image and the generated face.
Subsequently, we calculate the necessary affine transformation matrix $M$ that would align the face landmarks of the replacement ID perfectly with those of the original template. 
We then apply this matrix $M$ directly to warp the face landmarks of the generated face and substitute these into the original template.
As such, the warped face landmarks achieve both facial structure preservation and overall realism, resolving the issue of face landmark mismatch between the template and the specified ID.

\subsubsection{Second Inpainting Inference}
After generating the face landmarks of the specified ID, we perform the second inpainting inference to redraw the face region of the input template.
We first adopt the M2FP human parsing model to generate the mask of the face region and appropriately expand it to handle changes in facial shape.
To address the boundary artifacts arising from mask inpainting, we propose a novel approach for artifact rectification by combining multiple controls in the inpainting process.
The overall controls include:
(i) OpenPose-ControlNet using the warped face landmarks for facial structure preservation.
(ii) Canny-ControlNet using Canny edges of the areas outside the face of the input template to maintain harmonious edges.
(iii) Image latent of the input template by using a considerable inpainting strength (\eg 0.65).
As such, we redraw the face region of the template image to the specified ID, maintaining both identity preservation and overall realism.
It should be emphasized that we perform face fusion for the generated faces and portraits based on template face selection in both two inference procedures to improve facial details, which is elaborated in Section~\ref{sec:model-postpro}.

\subsubsection{Multiple User IDs}
FaceChain also supports the portrait generation of multiple user IDs by adjusting the regions for each face to perform the proposed inpainting pipeline.
Compared to single user ID portrait generation, the templates of multiple persons usually have higher resolution, and each face accounts for a smaller proportion of the entire image.
Thus, it is unnecessary and unsuitable to feed the entire image into the Stable Diffusion model for both text-to-image and inpainting processes.
Therefore, we crop the area around each face using DamoFD~\cite{liu2022damofd}, and perform both processes using the cropped image.
Then the generated portraits are merged back into the template.
However, there exists color difference at the edges of the cropped region due to the reconstruction error of the AutoEncoder of Stable Diffusion model.
Note that such color difference is independent of the diffusion process, since the image latent of the unmasked region is unchanged in inpainting.
As such, we add the reconstruction error back to the template and generate high-quality portraits for multiple user IDs.

\begin{figure*}[t]
    \centering
    \includegraphics[width=0.95\linewidth]{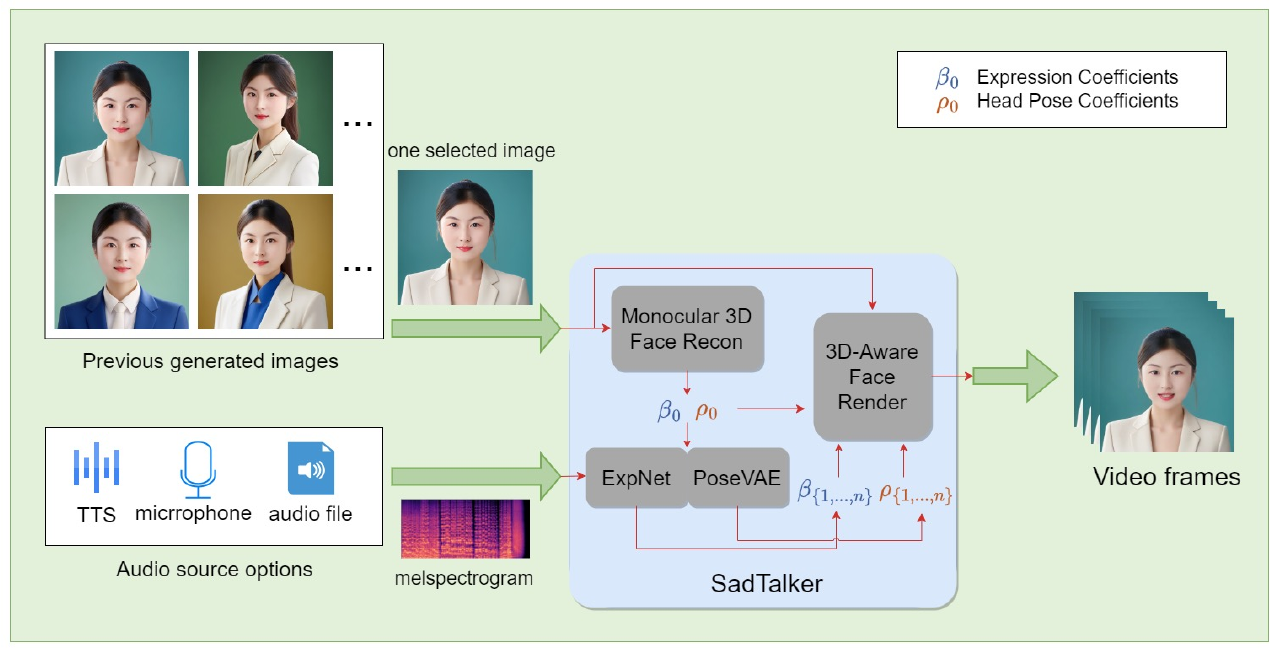}
    \caption{\textbf{Inference pipeline of talking head.} 
    SadTalker uses the coefficients of 3DMM as intermediate motion representation. So given single image and audio, Monocular 3D Face Reconstruction, PoseVAE and ExpNet will use them to generate realistic 3D motion coefficients (facial expression $\beta$, head pose $\rho$), then these coefficients are used to implicitly modulate the 3D-aware face render for final video generation.
    }
    \label{fig:sadtalker-inference}
\end{figure*}
\subsection{Infinite Style}
\label{sec:infinite-style}
Facilitating the generation of personalized portraits with diverse and infinite stylistic variations is the aim of FaceChain. To realize this vision, we have implemented the following: 
(i) We integrate multiple high-quality style-LoRA models, enabling the generation of portraits in an extensive range of styles.
(ii) We develop a plug-and-play module designed for the seamless incorporation of new style-LoRA models, ensuring convenient expandability.
Specifically, we have collected various of base models and style-LoRA models trained on specific styles from AI art generation communities (\eg, Civitai~\url{https://civitai.com/}). Furthermore, since the open-sourcing of FaceChain, many developers have voluntarily contributed their own trained style-LoRA models to the project. These models are integrated into ModelScope for convenient accessibility during the inference phase. 
To further enhance flexibility, we have developed an extensible plug-and-play module for integrating the local style-LoRA models. This module empowers users to leverage their own specified models during the inference phase, facilitating the generation of personalized portraits with highly customized stylistic variations.

Through the integration of models from developers and the AI art community, FaceChain presently provides access to nearly a hundred style-LoRA models. These models encompass a wide spectrum of artistic styles, enabling users to effortlessly generate personalized portraits in diverse styles.
From classy identification photos to futuristic cyberpunk-inspired art, FaceChain presents a myriad of creative possibilities, which can be seen in Fig. ~\ref{fig:infinite_style_1} and Fig. ~\ref{fig:infinite_style_2}. 
The plug-and-play style-LoRA model integration module allows users to seamlessly integrate their own trained models without modifying the FaceChain codebase. Additionally, users can contribute high-quality style-LoRA models to ModelScope, further enriching the project. 
This unique combination of an expansive style-LoRA model repository and user-friendly extensibility is designed to inspire infinite stylistic variations for personalized portraits.

\section{FaceChain Application}
\label{sec:application}
In this section, we further introduce two applications within the proposed FaceChain framework: Virtual Try-on and Talking Head.

\subsection{Virtual Try-on}
With the rapid progress of online shopping, there is an exploded demand for virtual try-on which provides virtual fitting experiences for specified person and garment.
According to whether person or garment information is fixed in the interactive experience, the core idea for virtual try-on methods can be categorized into generating digital twins of garments and persons.
As a playground for Face/Human-Centric identity-preserving portrait generation, FaceChain also supports virtual try-on with fixed garment and generated person.
Specifically, given a real or virtual model image with the garment to be tried on, as well as the trained face-LoRA model, FaceChain can redraw the region outside the garment including the person and backgrounds to get the virtual try-on result.

Similar to Section~\ref{sec:inpainting}, the redrawing process is also based on the inpainting procedure of the Stable Diffusion model.
However, we use different spatial conditional controls due to different requirements for the two tasks.
First, since all region outside the garment instead of merely the face region needs to be redrawn, the interaction between the face landmarks and the retained part is far lower.
Besides, the image contents of areas outside the garment may be harmful to image generation especially when we try to change the virtual model of the input template to a real model.
Therefore, we do not use the warped face landmarks in OpenPose-ControlNet, and adjust the inpainting strength to 1.0 to avoid the effect of the image latent of the input template.
Then, it is extremely important to generate portraits with harmonious body poses in virtual try-on, especially the detailed pose for hands, since most existing Stable Diffusion models cannot generate satisfying hands without fine control.
To address the above issue, we adopt the following controls to improve the quality of body pose:
(i) OpenPose-ControlNet using both bone and hand poses extracted by Distillation for Whole-body Pose (DWPose) estimators~\cite{yang2023effective} for more accurate hand gestures.
(ii) Depth-ControlNet using depth estimation~\cite{ranftl2020towards} on the hand region for 3D details of hands.
(iii) Canny-ControlNet using Canny edges of the body area for more harmonious body poses.
As for the background, we use prompt words to control it.
Finally, the generated virtual try-on images can be put into the inpainting pipeline to further improve the identity-preserving effectiveness.

\subsection{Talking Head}

In order to further explore the use case of the generated portrait, we thought of making the portrait alive, specifically, as illustrated in Fig.~\ref{fig:sadtalker-inference}, the user selects one face image from the previous generation, and then the user provides an audio clip, FaceChain will use them to output a video of speaking portrait. To achieve this, we tested several talking head/face algorithms~\cite{prajwal2020lip,zhang2023sadtalker,cheng2022videoretalking}, which can generate portrait video whose lips are corresponding to input audio. Finally, we chose sadtalker~\cite{zhang2023sadtalker} as our talking head backend algorithm, because wav2lip~\cite{prajwal2020lip} and video-retalking~\cite{cheng2022videoretalking} can only change the lips in the face, while PoseVAE and ExpNet proposed by SadTalker can change the head pose and facial expression respectively, in which the head pose can be selected from one of 46 embeddings, the expression coefficient can be controlled, and the eyes blink frequency can be controlled as well.

Because SadTalker's input and output resolutions are 256 or 512, if the face resolution in the original input image is relatively large, or if the user simply wants to make the overall resolution higher, we also support using GFPGAN~\cite{wang2021towards} as a post-processing module, which can double the resolution of the video. As for the audio clip, we support three options to provide it: (i) TTS synthesis, (ii) microphone recording, and (iii) local file uploading. As for TTS synthesis, we use the API from Microsoft Edge, because it supports multi-language mixed input, and supports multiple languages synthesis, which make FaceChain more friendly to users around the world.

\section{Future Work}
\label{sec:future work} 
Given multiple images depicting the same individual, FaceChain can generate a diverse collection of high-fidelity, identity-preserving portraits with distinct stylistic variations.
These variations encompass a wide spectrum, ranging from classy identification photos and human portraits, to photos of futuristic aesthetics of the cyberpunk genre. 
Still, we acknowledge current work on FaceChain is merely scratching the tip of the iceberg, and an immerse universe of applications along the line is waiting to be explored. In this Section, we put forward a few directions we consider worthwhile to explore.
\begin{itemize}
    \item Personalized generation framework capable of handling multiple subjects of different ages and genders.
    \item Improved data processing mechanism to retain stature impeccably, which will require more diverse training data.
    \item Support adaptive weight-selection for style and face LoRA models during model fusion process.
    \item Encode diverse styles information into a unified model that can be activated with specific triggering prompts.
    \item Develop tailored similarity ranking and face Fusion models for FaceChain.
    \item Explore train-free framework for customized portrait generation. Current approach used in FaceChain requires a new model to be trained for each human id, which can be  computational expensive for wide adoption.
\end{itemize}

\noindent\textbf{Acknowledgments.} We wish to thank Dr. Peiran He and  Miaomiao Cui
for many helpful discussions.

\newpage

{\small
\bibliographystyle{ieee_fullname}
\bibliography{egbib}
}

\end{document}